\documentclass[runningheads]{llncs}
\usepackage{amsmath}
\usepackage{graphicx}
\usepackage{caption}
\usepackage{bbm}
\usepackage{subfig}
\usepackage{tabularx}
\usepackage{multirow}

\usepackage[ruled,vlined, linesnumbered]{algorithm2e}
\SetKwInput{KwInput}{Input}                
\SetKwInput{KwOutput}{Output}              

\newcommand{\bftab}{\fontseries{b}\selectfont}
\newcolumntype{C}[1]{>{\centering\arraybackslash}p{#1}}

\begin{document}
\title{Improving Sample Efficiency with Normalized RBF Kernels}
%
%
\author{Sebastian Pineda-Arango \and David Obando-Paniagua \and Alperen Dedeoglu \and Philip Kurzendörfer \and Friedemann Schestag \and Randolf Scholz}
\authorrunning{S. Pineda-Arango et al.}
%
\institute{University of Hildesheim, Samelsonplatz 22, Hildesheim 31141, Germany \\
\email{\{pineda, obandod, dedeoglu, kurzendo, schestag\}@uni-hildesheim.de, rscholz@ismll.uni-hildesheim.de}\\}
\maketitle              
\begin{abstract}
In deep learning models, learning more with less data is becoming more important. This paper explores how neural networks with normalized Radial Basis Function (RBF) kernels can be trained to achieve better sample efficiency. Moreover, we show how this kind of output layer can find embedding spaces where the classes are compact and well-separated. In order to achieve this, we propose a two-phase method to train those type of neural networks on classification tasks.  Experiments on CIFAR-10 and CIFAR-100 show that networks with normalized kernels as output layer can achieve higher sample efficiency, high compactness and well-separability through the presented method in comparison to networks with SoftMax output layer.

\end{abstract}

\section{Introduction}
Poor sample efficiency represents a challenge in current machine learning models. While humans are excellent at learning from very few samples, deep learning models are not generally sample efficient and require a lot of training data to achieve good performance. For this reason, in the last years there has been an increasing interest in how to construct models that generalize while only having few samples on training, also called few-shot learning. In this area, some prominent models have been proposed such as Prototypical Neural Networks \cite{snell2017prototypical}  or Matching Networks \cite{Vinyals16}.

Sample efficient models are especially interesting in reinforcement learning settings, since it means less training epochs to achieve a given performance level. This was the main objective of Pritzel et al. in \cite{Pritzel}. They used a differential neural dictionary to store latent state representations and the corresponding Q-values as key-value pairs and read the dictionary using a 
kernel smoothing model with a radial basis function kernel. The dictionary was basically a key-value structure that can be read through a kernel smoothing model with a Radial Basis Function (RBF) kernel (inverse kernel).

Given the positive results presented in Neural Episodic Control (NEC) for sample efficiency through a kernel output layer, this paper explores the idea of kernel smoothing, specifically using RBF, in classification problems as a way of improving sample efficiency. \cite{Lippmann,Park}. Additionally, RBF networks are  interesting to explore, since they enable robust classification \cite{Goodfellow}, rejection of unknown classes \cite{Zadeh} and few-shot learning \cite{snell2017prototypical}.

We propose a training process that consists of a pre-training phase using a SoftMax output, after which the centers of the RBF kernel are initialized using \textit{K-Means}, to ensure optimal separability. Subsequently, it is trained with RBF kernel as output. Our hypothesis is that by using an output layer that maximizes the inter-class separability and reduces the intra-class compactness, as RBF networks do, it is possible to improve sample efficiency. Therefore, finding a good training procedure for these type of networks is important. 

Experiments on CIFAR-10 and CIFAR-100 \footnote{Code is available at https://github.com/anonym-submission-2020/n-rbf-kernels} show that kernel with pre-training and \textit{K-Means} initialization is able to achieve good sample efficiency, especially when using only small percentages of the available training set. These elements also allow the network with a kernel output layer to achieve better accuracy compared to training without any kind of initialization, in spite of the well-known difficulty of training RBF-networks \cite{Goodfellow}.

The rest of this paper is organized as follows. In Section \ref{background}, we present the previous work. In Section \ref{Model}, we introduce the mathematical formulation and the two-phase training procedure. We provide the experimental evaluation and the results in Section \ref{Experiments} and conclude our work in Section \ref{Conclusion}.

\section{Background and related work} \label{background}

The idea of using kernelized output layers is closely related to several type of tasks such as metric learning and few shot learning. Moreover, their use in neural networks architectures resembles RBF networks \cite{Park} and neural networks with memory modules \cite{Kaiser,Orhan,Sprechmann}. Neural Networks using RBF functions were introduced by Powell \cite{Powell} to deal with the interpolation problem in a multidimensional space\cite{Powell} and they have been traditionally trained in ways that include different phases \cite{Schwenker}. The first deep neural network using a RBF output layer, called LeNet5, was introduced by LeCun et al. \cite{LeCun}. Later, Simard et al. \cite{Simard} suggested that using a SoftMax output layer instead can achieve slightly  higher accuracy ad could be optimized faster. Afterwards, SoftMax output layer became the standard.

J. Xu et al. \cite{Zhang} proposed a kernel neuron, as an alternative to the classical neuron, and which could be trained through gradient optimization algorithms. However, it did not impact the further develop of neural networks, which kept using neurons other activation functions and SoftMax output layer. The kernels and the distance metrics were still used in the field of metric learning.  In this context, Jacob Goldberger et al. \cite{Goldberger} introduced Neighbourhood Components Analysis, a method for learning Mahalanobis distance measure to be used in the KNN classification by using a SoftMax over euclidean distances. With success of deep learning, new architectures were proposed for the distance metric task in the frame of deep metric learning to learn distance metrics using deep neural networks. For instance, Benjamin J. Meyer et al. \cite{Meyer} , for instance, used deep metric learning with nearest neighbour gaussian kernels as output layers.

In recent years, RBF output networks are starting to be appealing again thanks to their non-linearity and robustness. Prototypical networks \cite{snell2017prototypical} are neural architectures that enable few-shot learning by learning an embedding space where samples of the same class are localized around the prototype. In here, they use RBF to associate embeddings to their prototypes. Besides this, the work of Orhan in \cite{Orhan} presented how RBF in the prediction with memory modules are an opportunity to deal with catastrophic forgetting. On the other hand, Qi Qian et al. \cite{Qian} proposed the SoftTriple Loss, a cost function that enables a sample efficient deep metric learning through \textit{Relaxed Similarity}. However, they did not experiment with RBF kernels.

\section{Model} \label{Model}

\subsection{Network with general output layer} \label{GeneralOutput}

\begin{figure}
    \centering
    \includegraphics[scale=0.25, trim={0 6cm 0 6cm}, clip]{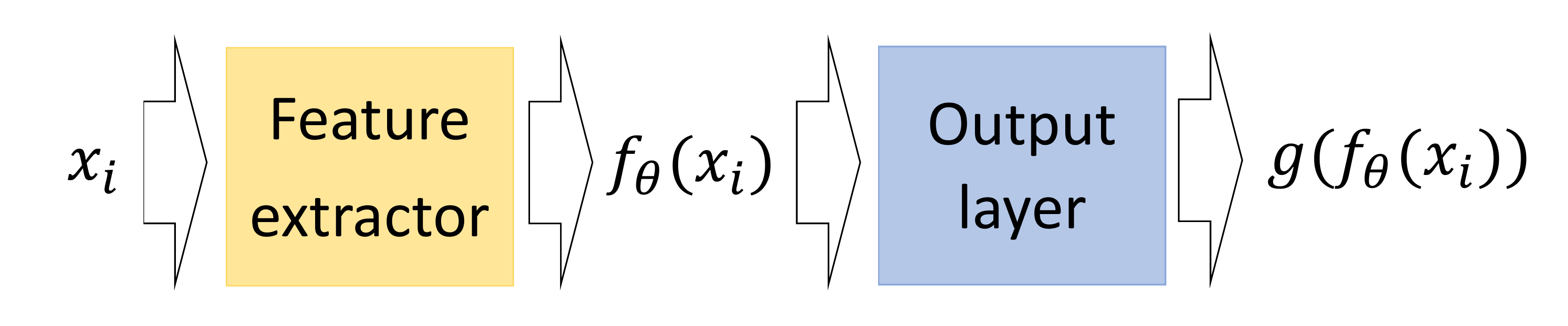}
    \caption{General output layer.}
    \label{fig:my_label}
\end{figure}

In this section, an abstraction of a general neural architecture is presented, and it will be used in further references throughout the paper. Given an input $\textbf{x}_i$, a neural network architecture with parameters $\boldsymbol{\theta}$ can be simplified as a feature extractor whose output, $f_\theta(\textbf{x}_i) \in {\rm I\!R}^P$ is an embedding of the input. Afterwards, the embeddings are the input of a final output layer $g(\cdot)$, $g:{\rm I\!R}^P  \rightarrow {\rm I\!R}^C $ , which outputs the probability of belonging to a class. Therefore, in general, the probability output of this architecture in a classification task corresponds mathematically to:

\begin{equation}
Pr(Y=c |\textbf{x}_i) = (g(f_\theta(\textbf{x}_i))_c 
\label{eq:1}
\end{equation}

\subsection{SoftMax output}

SoftMax as an output layer has been the default option in many of the deep convolutional neural architectures after the first introduction by Simard et al. \cite{Simard}. Following the notation introduced in Section \ref{GeneralOutput}, the SoftMax output layer, including the output weights $[\textbf{w}_1, ...,\textbf{w}_C] \in  {\rm I\!R}^{P\times C}$, can be written as:

\begin{equation}
Pr(Y=c |\textbf{x}_i) =  \frac{\rm exp(\textbf{w}_c^T \textit{f}_\theta(
\textbf{x}_i))}
{\sum_{j=1}^C{\mathrm {exp}(\textbf{w}_j^T \textit{f}_\theta(\textbf{x}_i))}}
\label{eq:2}
\end{equation}
Additionally, the cross-entropy loss can be defined for the SoftMax by denoting $y_i \in \{1, ..., C\}$ as the label of $\textbf{x}_i$.
\begin{equation}
\mathcal{L}(\textbf{x}_i, y_i,\boldsymbol{\theta}, \textbf{w})_{\text {SoftMax}}= - \text{log} \frac{\text{ exp}(\textbf{w}_{y_i}^T \textit{f}_\theta(
\textbf{x}_i))}
{\sum_{j=1}^C{\text {exp}(\textbf{w}_j^T \textit{f}_\theta(\textbf{x}_i))}}
\label{eq:3}
\end{equation}
The loss in the equation \ref{eq:3} minimize the angle between $\textbf{w}_{y_i}$ and $f_\theta(\textbf{x}_i)$, if the magnitude of them are controlled (through regularization and batch normalization, respectively). It means, that by using SoftMax, the optimization of the respective loss (equation \ref{eq:3}) aims to find well-separated classes. 

\subsection{RBF kernel output}

Another possible output function is a normalized RBF kernel output. This has been original formulated for regression problems (\textit{Nadaraya-Watson model} or \textit{kernel smoothing} \cite{Nadaraya}). However, its formulation can be extended to a probabilistic output layer of a neural network which can be used for classification settings:

,\begin{equation}
Pr( Y=c |\textbf{x}_i)  = \frac{\sum^Q_{j=1}{v_{j,c}\kappa( f_\theta(\textbf{x}_i)-\boldsymbol{\mu}_j})}
				{\sum^Q_{j=1}{\kappa(f_\theta(\textbf{x}_i)-\boldsymbol{\mu}_j})} ,
\label{eq:6}
\end{equation}
where $\kappa(\cdot)$ is the kernel function, $\boldsymbol{\mu}_j$ is the j-th center and $Q$ is the total number of centers. Given that every center $\boldsymbol{\mu}_j$ is associated to only one class through a label $l_j$, the value $v_{j,c}$ can be computed as $\mathbbm{1}_{l_j =c}$. The formulation in equation \ref{eq:6} is similar to the reading mechanism in memory augmented networks \cite{Santoro}. However, it is possible to define it in a more compact way by defining, $K_c = \{\boldsymbol{\mu}_j | l_j =c\}$ the set of all centers belonging to a given class $ c$:

\begin{equation}
Pr(Y=c |\textbf{x}_i)  = \frac{\sum_{\boldsymbol{\mu}_j \in K_c}{\kappa(f_\theta(\textbf{x}_i)-\boldsymbol{\mu}_j})}
				{\sum^Q_{j=1}{\kappa(f_\theta(\textbf{x}_i)-\boldsymbol{\mu}_j})} 
\label{eq:7}
\end{equation}
The cross-entropy applied over equation \ref{eq:7} yields a formulation similar to equation \ref{eq:3}:

\begin{equation}
\mathcal{L}(\textbf{x}_i, y_i, \boldsymbol{\theta}, \boldsymbol{\mu})_{\rm Kernel}  = - \text {log} \frac{\sum_{\boldsymbol{\mu}_j \in\textit{K}_{y_i}}{\kappa(\textit{f}_\theta(\textbf{x}_i)-\boldsymbol{\mu}_\textit{j}})}
				{\sum^Q_{j=1}{\kappa(\textit{f}_\theta(\textbf{x}_i)-\boldsymbol{\mu}_j})} 
\label{eq:8}
\end{equation}

In this work, only the Gaussian kernel, $\kappa(\textbf{x}- \boldsymbol{\mu}) = \rm exp(-\alpha||\textbf{x}-\boldsymbol{\mu}||^2)$, and the inverse distance kernel, $\kappa(\textbf{x}- \boldsymbol{\mu}) = \frac{1}{||\textbf{x}-\boldsymbol{\mu}||_2^2+\delta}$ , are considered. 
In the equation \ref{eq:8}, the centers are the parameters to be learnt. Although the number of center per class, $k=|K_c|$, can be greater than one, this increases the complexity of the model. 
When having only one center per class and the kernel is Gaussian, it is possible to reinterpret the RBF output as a SoftMax output,

\begin{equation}
Pr(Y=c |\textbf{x}_i)  = \frac{{\text {exp} (-\alpha||\textit{f}_\theta(\textbf{x}_i)-\boldsymbol{\mu}_c}||^2)}
				{\sum^Q_{j=1}{\text {exp} (-\alpha||\textit{f}_\theta(\textbf{x}_i)-\boldsymbol{\mu}_j}||^2)} \\
                \\
            =    \frac{\text {exp} (\textbf{w}_c^T \textit{f}_\theta(\textbf{x}_i) + \textit{b}_c)}
				{\sum^Q_{j=1}{\text {exp} (\textbf{w}_j^T \textit{f}_\theta(\textbf{x}_i) + \textit{b}_j)} }
\label{eq:9}
\end{equation}
where $\textbf{w}_j=2\alpha\boldsymbol{\mu}_j$ and $b_j = -\alpha\boldsymbol{\mu}^T_j\boldsymbol{\mu}_j$. Hereby. we expanded the term in the exponent that corresponds to the squared euclidean distance \cite{snell2017prototypical}.  However, the key difference is that the loss function obtained after applying the output expressed in equation \ref{eq:9} into equation \ref{eq:8} enables the learning of embedding spaces that maximize the inter-class separability and minimize the intra-class compactness. 

An advantage of the output function of equation \ref{eq:7} is that there are not further parameters to train besides the centers of the kernels, whereas the literature traditionally propose weights over the kernels. Therefore, it is possible to interpret directly the final centers as class prototypes.

\subsection{Training procedure of networks with RBF Kernel Output  }

As discussed in the previous section, the RBF kernel output with a cross-entropy loss function achieves explicitly inter-class separability and compactness. However, these RBF units have been receiving few attention in the main deep learning models, and, in some cases, they are regarded as having less accuracy in training \cite{Goodfellow}. Therefore, to refine the training, we propose a two phases procedure as follows:

\begin{enumerate}
    \item \textbf{Phase 1 (P1)}. In this phase, a network with feature extractor $f_{\theta}(\cdot)$ plus a classical SoftMax output layer is trained. Here, it refers also to the weights of the fully connected layer before the SoftMax, \textbf{w}, as expressed in equation \ref{eq:2}. This network is trained by optimizing the loss function in equation \ref{eq:3} until convergence using the training set $\mathcal{D}_{Train} = \{(\textbf{x}_1, y_1), ..., ( {\textbf{x}_N, y_N})$ as in the equation \ref{Phase1}.\\

\begin{equation} \label{Phase1}
\boldsymbol{\theta}, \textbf{w} = \mathrm{arg min}_{\boldsymbol{\theta}, \textbf{w}} \sum_{(\textbf{x}_i, y_i) \in \mathcal{D}_{Train}}\mathcal{L}(\textbf{x}_i, y_i,\boldsymbol{\theta}, \textbf{w})_{\text {SoftMax}}
\end{equation}
    \item \textbf{Initialization (Init)}. The previously trained network is used to generate embeddings $\textbf{\~{x}} \in {\rm I\!R}^P$ of all the training samples \textbf{x}. For each class $c$, we define $\mathcal{D}_c$, which is the subset of $\mathcal{D}_{train}$ containing all elements ($\textbf{x}_i, y_i$) such that $ y_i = c $. These elements are clustered in $k$groups using \textit{K-Means} algorithm \cite{MacQueen}. After clustering, the final set of centers, $K_c$ is saved for each class. The initialization procedure is summarized in Algorithm \ref{alg:RBFlearning}.\\

    \item \textbf{Phase 2 (P2)}. In this last stage, the parameters of the previously trained feature extractor are kept, while the output layer is changed. The set of centers $K_c$ are used as initial parameters (centers) for a RBF Kernel output layer (equation 2). The network is trained until convergence by optimizing the loss function in equation \ref{eq:8} and which is described by the equation \ref{Phase2}. 
\end{enumerate}
\begin{equation} \label{Phase2}
\boldsymbol{\theta}, \boldsymbol{\mu} = \mathrm{arg min}_{\boldsymbol{\theta}, \boldsymbol{\mu}} \sum_{(\textbf{x}_i, y_i) \in \mathcal{D}_{Train}}\mathcal{L}(\textbf{x}_i, y_i,\boldsymbol{\theta}, \boldsymbol{\mu})_{\text {Kernel}}
\end{equation}

\begin{figure}[!tbp]
  \centering
  \begin{minipage}[b]{0.3\textwidth}
    \subfloat[Phase 1 (P1)]{\includegraphics[scale=0.2, trim={6cm 0cm 6cm 0cm}, clip]{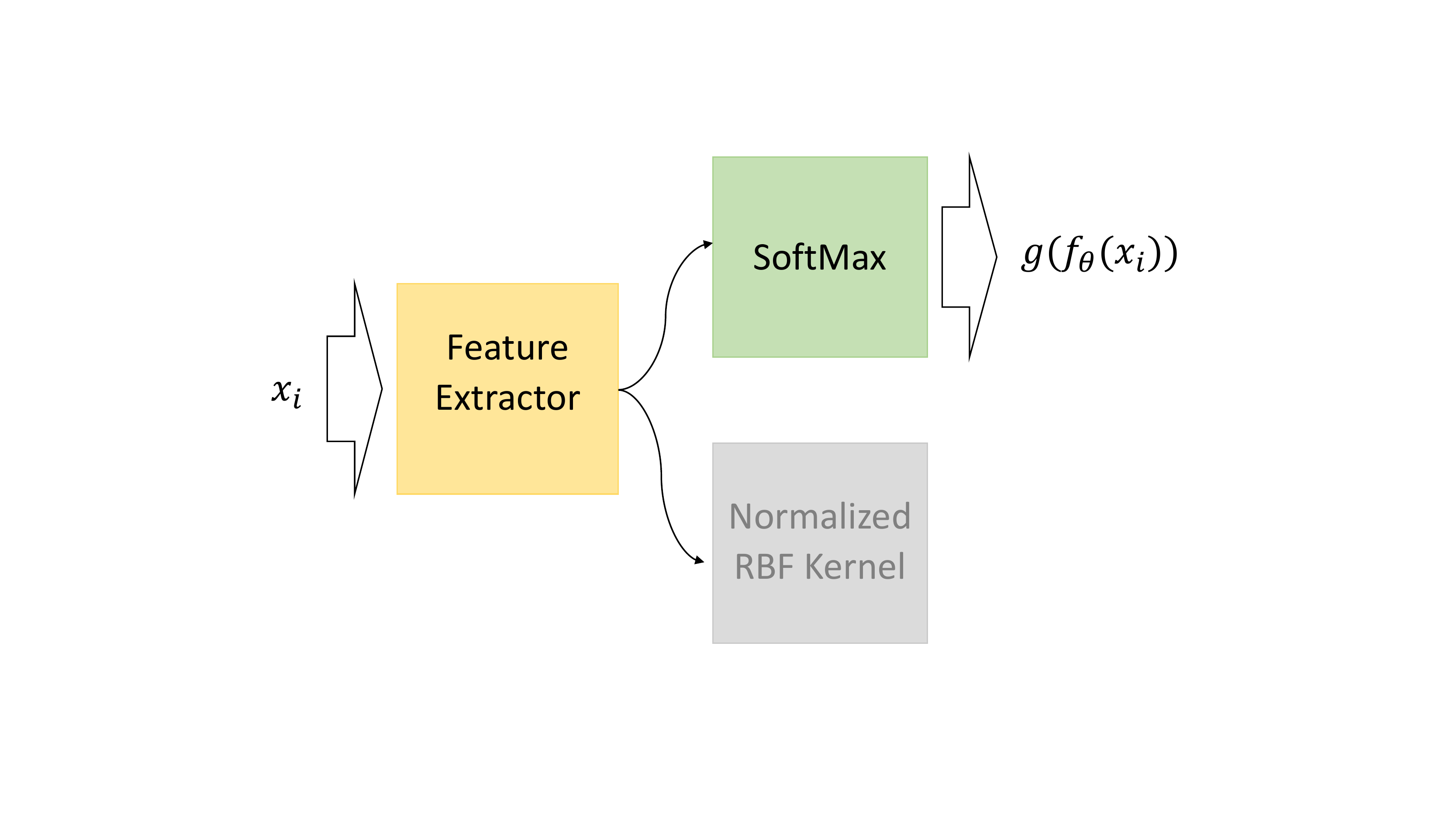}}
  \end{minipage}
  \hfill
  \begin{minipage}[b]{0.3\textwidth}
    \subfloat[Initialization (Init)]{\includegraphics[scale=0.2, trim={6cm 0cm 6cm 0cm}, clip]{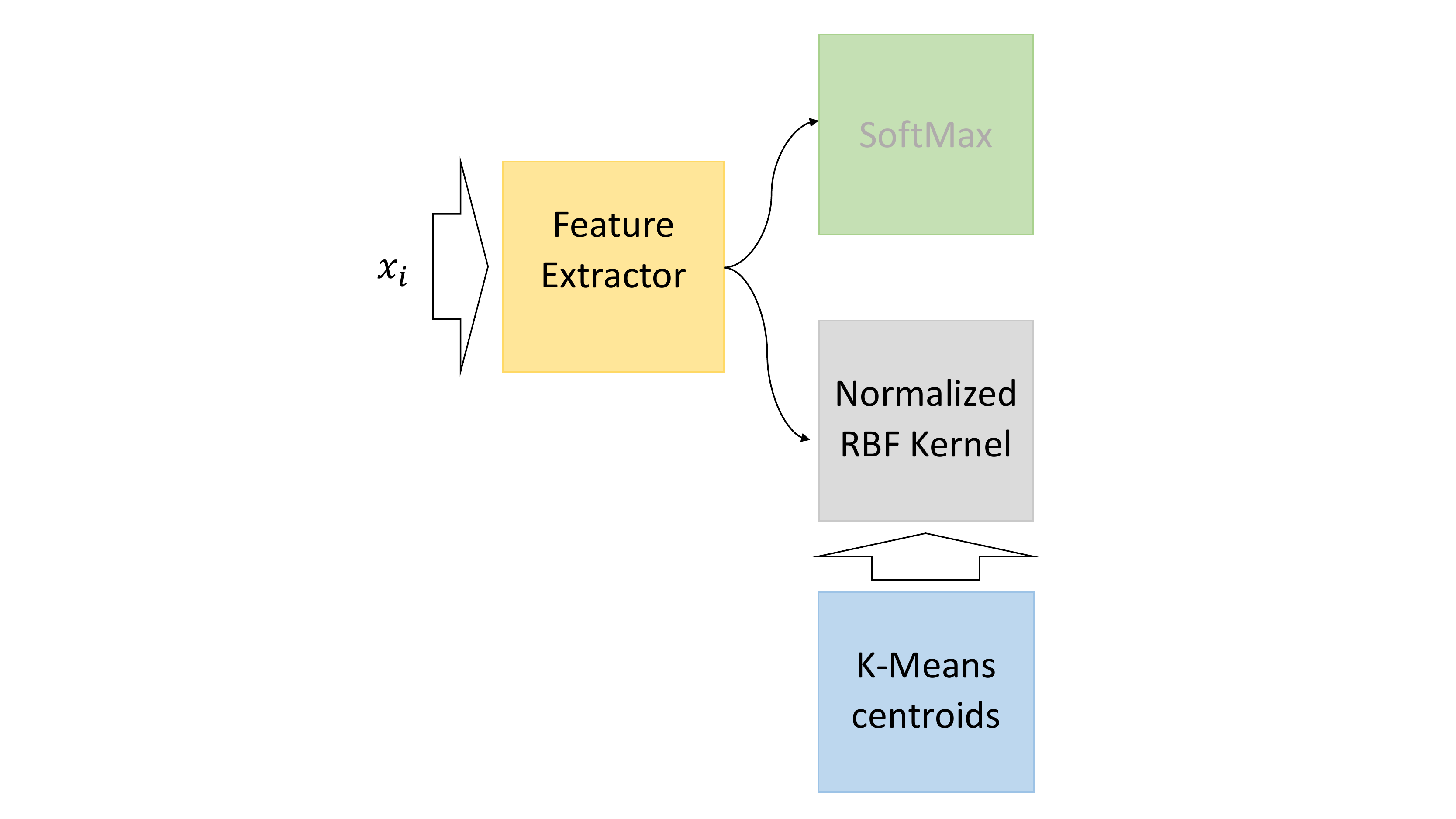}}
  \end{minipage}
  \hfill
  \begin{minipage}[b]{0.3\textwidth}
    \subfloat[Phase 2 (P2)]{\includegraphics[scale=0.2, trim={6cm 0cm 6cm 0cm}, clip]{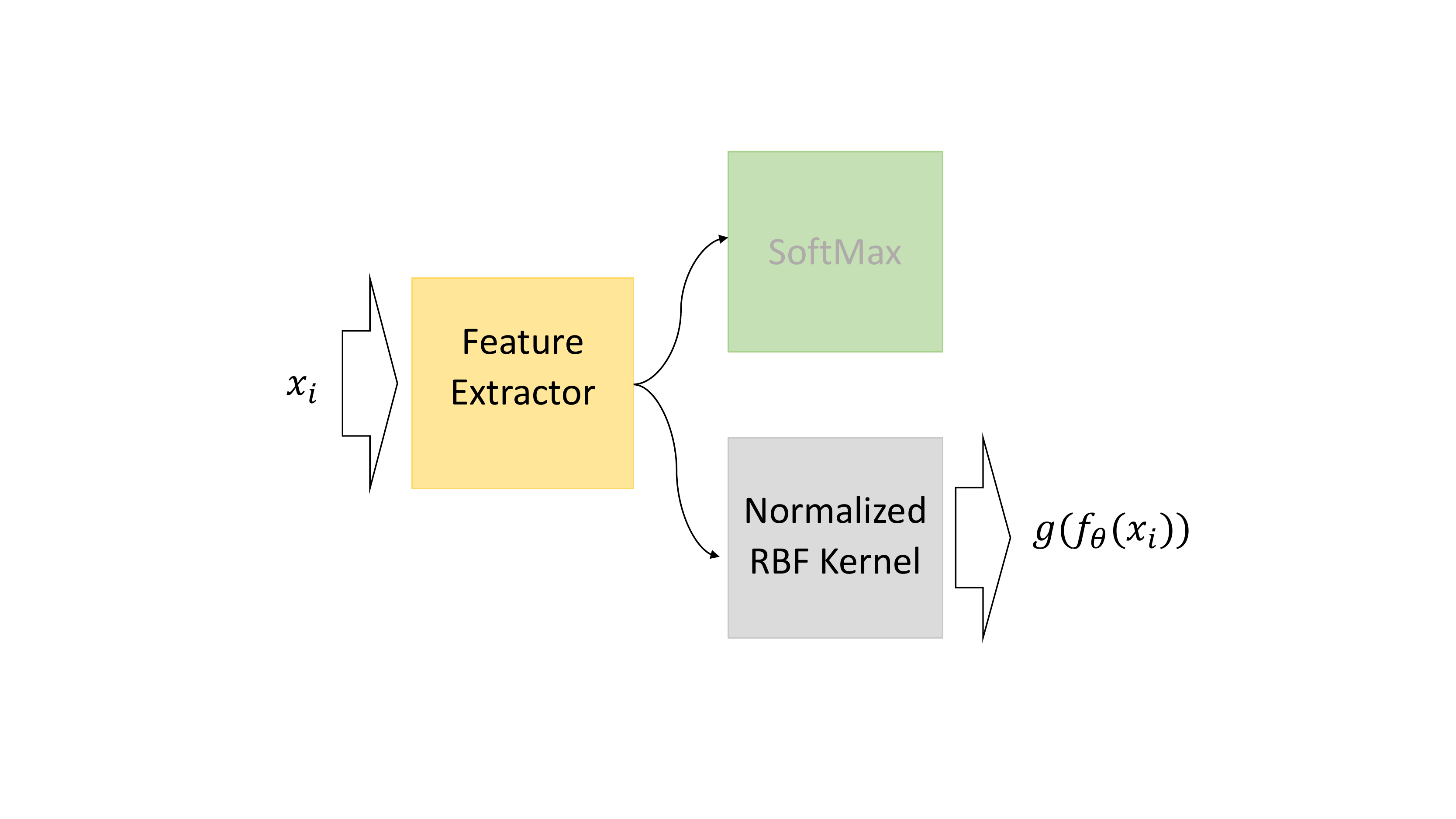}}
  \end{minipage}
  \caption{Training phases to train a neural network using Normalized RBF Kernels. First,(a) the neural network with SoftMax output is trained. Then, (b) the parameters of the new output layer are initialized. Finally, (c) the network with RBF output is trained. }
  \label{fig:RBFlearning}
\end{figure}

The general procedure to train the network is shown in Figure \ref{fig:RBFlearning}. 
\\

\begin{algorithm}[H]
\label{alg:RBFlearning}
\caption{Initialization algorithm }
 \KwInput{Parameters: $\boldsymbol{\theta, \textbf{w}}$
\newline
 Training set : $\mathcal{D}_{Train}$ = $\{(\textbf{x}_1, y_1), ..., ( {\textbf{x}_N, y_N})\}$. }

 \KwOutput{Trained parameters $\boldsymbol{\theta}, \boldsymbol{\mu}.$ }

\For{$c\leftarrow 1$ \KwTo $C$ }{

	 $\tilde{\textbf{x}_i}   \leftarrow  f_\theta(\textbf{x}_i), \space \space \textbf{x}_i \in \mathcal{D}_c$\;

	\text{clustering} $\leftarrow$ \textsc{kMeans($\tilde{\textbf{x}} , k$)}
	
	$K_c \leftarrow$ \text{\textsc{getCenters}(clustering)}
	}
\end{algorithm}

\section{Experiments} \label{Experiments}

\begin{figure}[ht]
\centering
\subfloat[CNN, CIFAR-10]{\label{fig:mdleft}{\includegraphics[width=0.5\textwidth]{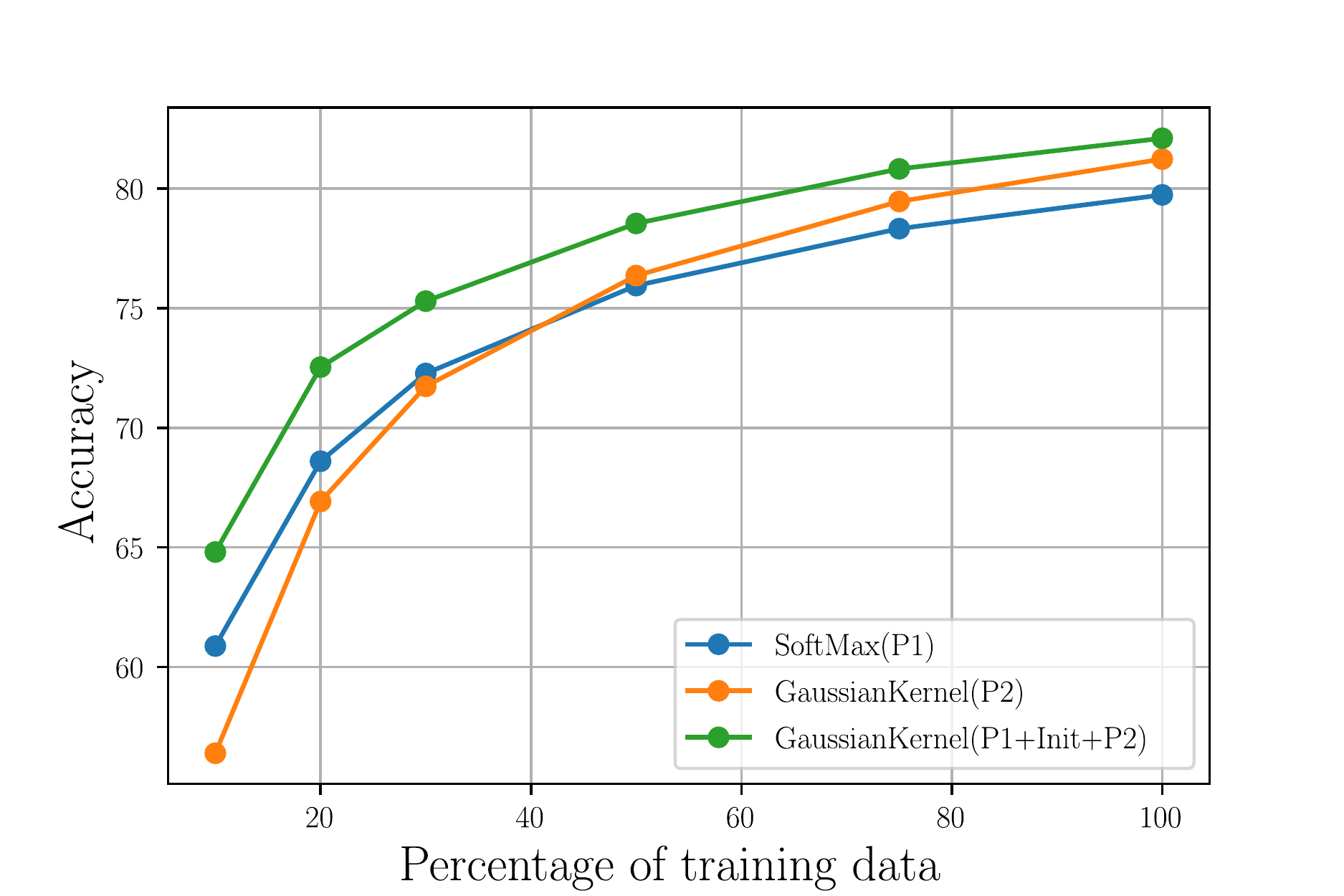}}}\hfill
\subfloat[CNN, CIFAR-100]{\label{fig:mdright}{\includegraphics[width=0.5\textwidth]{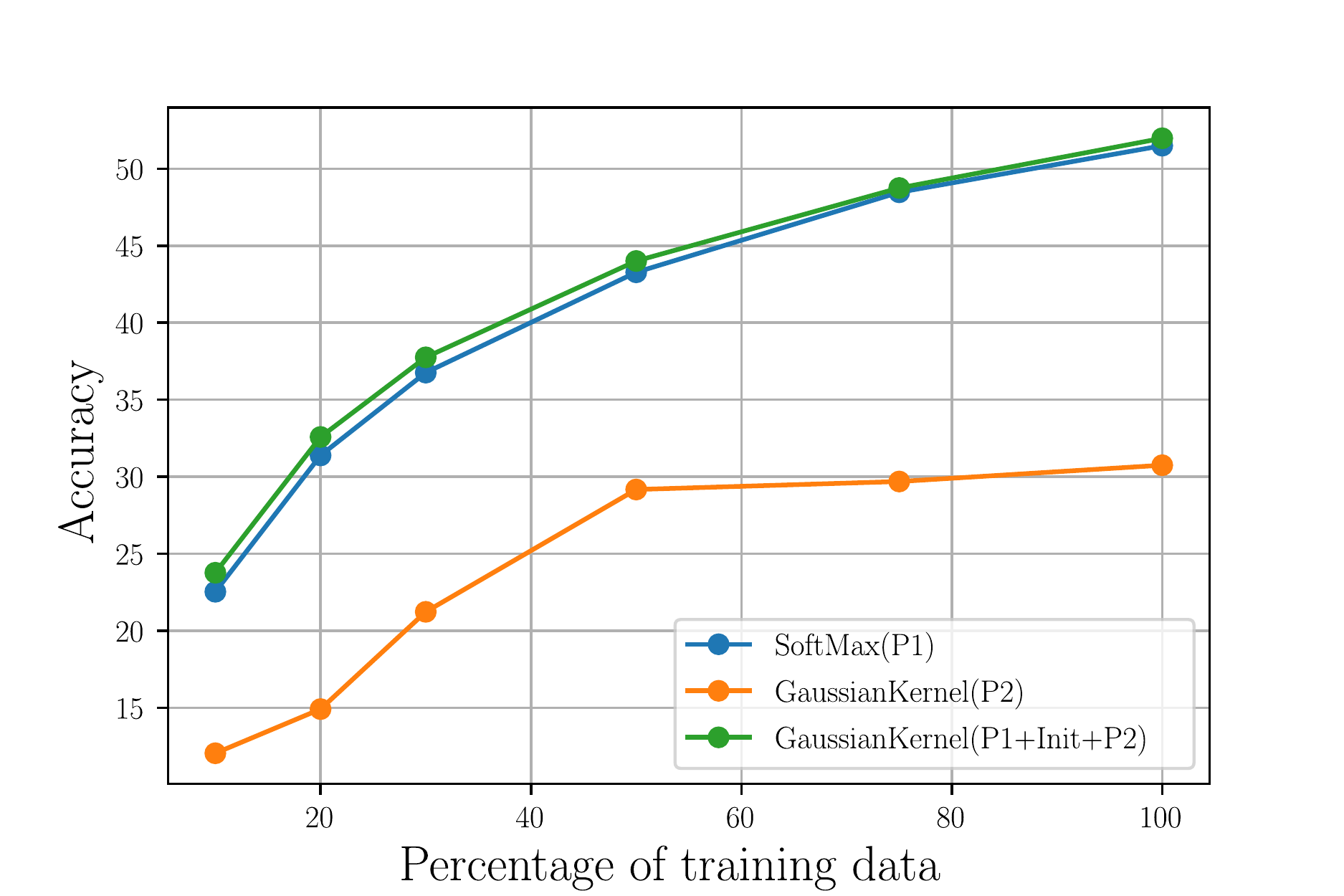}}}

\subfloat[ResNet50, CIFAR-10]{\label{fig:mdleft}{\includegraphics[width=0.5\textwidth]{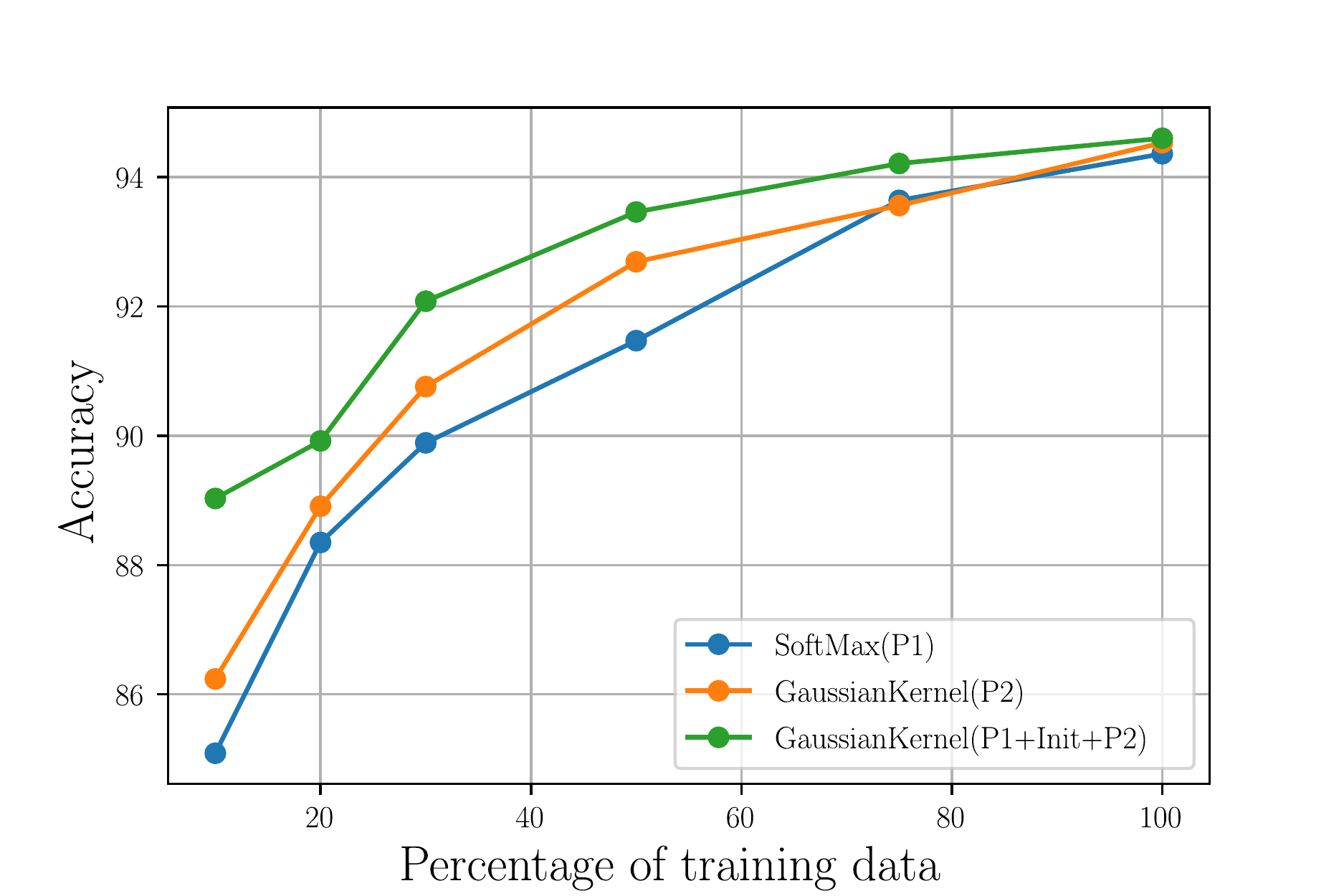}}}\hfill
\subfloat[ResNet50, CIFAR-100]{\label{fig:mdright}{\includegraphics[width=0.5\textwidth]{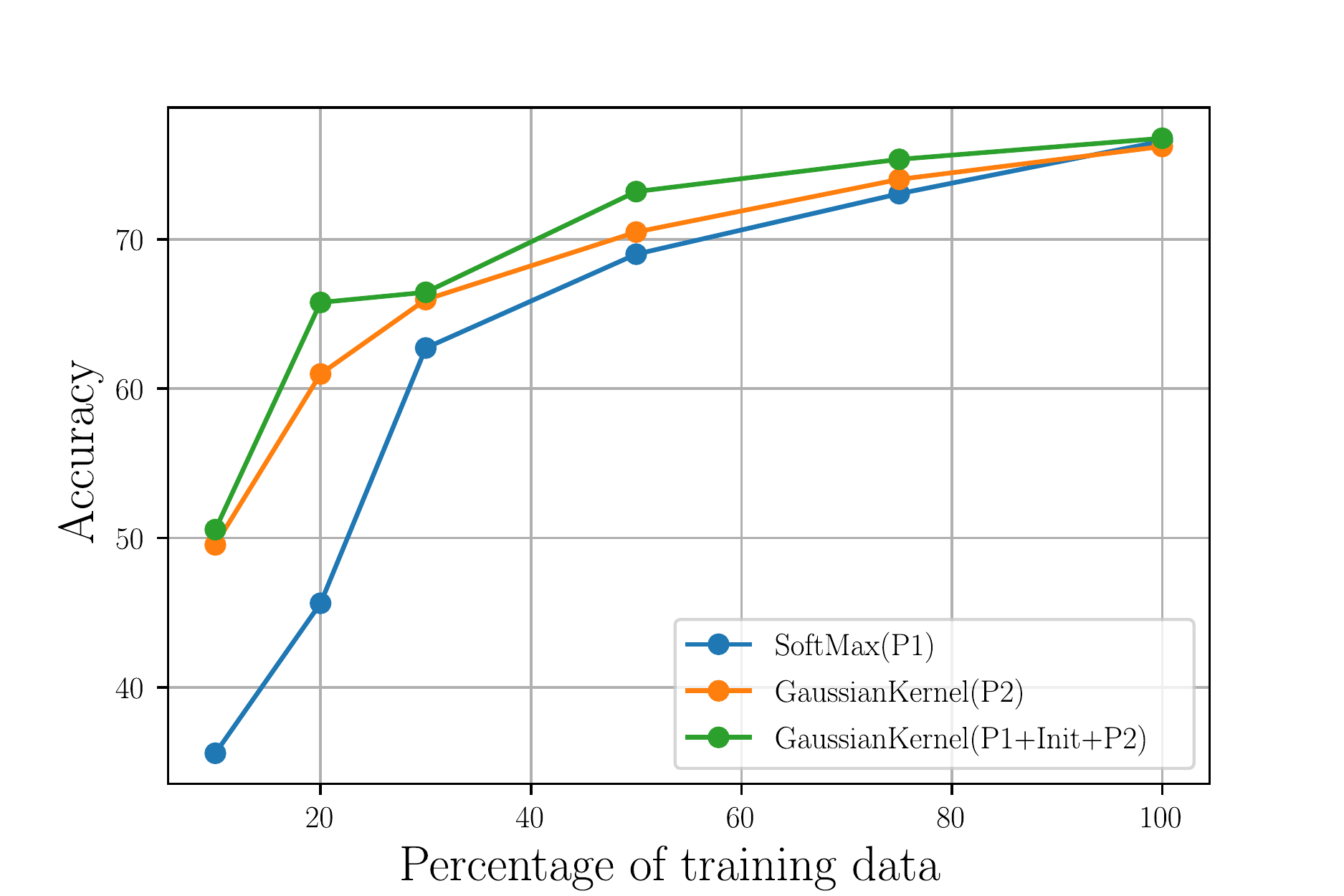}}}
\caption{Accuracy for different percentages of training data.}
\label{fig:accuracy-percentages}
\end{figure}

\begin{table}[h]
\caption{Results with CNN on CIFAR-10.}
\centering
\begin{tabular}{|c|c|c|c|c|c|c|}
\cline{2-7} \hline
                                        \textbf{Training}        & \textbf{10\%}  & \textbf{20\%}  & \textbf{30\%}  & \textbf{50\%}  & \textbf{75\%}  & \textbf{100\%} \\ \hline
\multicolumn{1}{|l|}{\text{SoftMax(P1)}}              & 60.88         & 68.60          & 72.27          & 75.94          & 78.32          & 79.73          \\ 
\multicolumn{1}{|l|}{\text{InverseKernel(P2)}}         & 58.75          & 61.25          & 68.98          & 71.40          & 75.40          & 77.69          \\ 
\multicolumn{1}{|l|}{\text{InverseKernel(P1+P2)}}    & 62.89          & 69.10          & 71.32          & 75.20          & 76.90          & 78.62          \\ 
\multicolumn{1}{|l|}{\text{InverseKernel(P1+Init)}}    & 60.01          & 67.15          & 71.10          & 73.93          & 75.93          & 76.98          \\ 
\multicolumn{1}{|l|}{\text{InverseKernel(P1+Init+P2)}} & 63.13          & 70.29          & 73.51          & 77.37          & 79.17          & 80.36          \\ 
\multicolumn{1}{|l|}{\text{InverseKenel(P1+Init2+P2)}} & 62.82          & 70.19          & 72.92          & 76.00          & 78.27          & 79.62          \\ 
\multicolumn{1}{|l|}{\text{RelaxedSimilarity(P2)}}     & 57.16          & 61.09          & 62.32          & 66.02          & 67.04          & 68.23          \\ 
\multicolumn{1}{|l|}{\text{GaussianKernel(P2)}}        & 56.40          & 66.92          & 71.73          & 76.36          & 79.46          & 81.23          \\ 
\multicolumn{1}{|l|}{\text{GaussianKernel(P1+Init+P2)}}     & \bftab{64.81} & \bftab{72.54} & \bftab{75.29} & \bftab{78.54} & \bftab{80.82} & \bftab{82.10} \\ \hline
\end{tabular}

\label{tab:results-cnn-c10}
\end{table}

\begin{table}[h]
\caption{Results with CNN on CIFAR-100.}
\centering
\begin{tabular}{|c|c|c|c|c|c|c|}
\cline{2-7} \hline
                  \textbf{Training}                              & \textbf{10\%}  & \textbf{20\%}  & \textbf{30\%}  & \textbf{50\%}  & \textbf{75\%}  & \textbf{100\%} \\ \hline
\multicolumn{1}{|l|}{\text{SoftMax(P1)}}              & 22.54          & 31.39          & 36.76          & 43.27          & 48.47          & 51.49          \\ 
\multicolumn{1}{|l|}{\text{InverseKernel(P2)}}         & 17.56          & 22.24          & 24.60          & 27.42          & 27.83          & 30.63          \\ 
\multicolumn{1}{|l|}{\text{InverseKernel (P1+P2)}}    & 19.20          & 24.32          & 25.71          & 27.41          & 29.06          & 28.18          \\ 
\multicolumn{1}{|l|}{\text{InverseKernel(P1+Init)}}    & 22.21          & 30.15          & 34.57          & 40.06          & 44.42          & 45.84          \\ 
\multicolumn{1}{|l|}{\text{InverseKernel(P1+Init+P2)}} & 21.71          & 29.36          & 33.38          & 38.06          & 42.73          & 44.18          \\ 
\multicolumn{1}{|l|}{\text{InverseKenel(P1+Init2+P2)}} & 11.87          & 20.19          & 25.18          & 33.44          & 33.44          & 34.83          \\ 
\multicolumn{1}{|l|}{\text{RelaxedSimilarity(P2)}}     & 17.75          & 19.72          & 20.98          & 21.09          & 20.80          & 21.66          \\ 
\multicolumn{1}{|l|}{\text{GaussianKernel(P2)}}        & 12.05          & 14.92          & 21.23          & 29.17          & 29.69          & 30.75          \\ 
\multicolumn{1}{|l|}{\text{GaussianKernel(P1+Init+P2)}}     & \bftab{23.77} & \bftab{32.58} & \bftab{37.75} & \bftab{44.01} & \bftab{48.75} & \bftab{51.97} \\ \hline
\end{tabular}

\label{tab:results-cnn-c100}
\end{table}

\begin{table}[h]
\caption{Results with ResNet50 on CIFAR-10.}
\centering
\begin{tabular}{|c|c|c|c|c|c|c|}
\cline{2-7} \hline
                                      \textbf{Training}           & \textbf{10\%}  & \textbf{20\%}  & \textbf{30\%}  & \textbf{50\%}  & \textbf{75\%}  & \textbf{100\%} \\ \hline
\multicolumn{1}{|l|}{\text{SoftMax(P1)}}                & 85.09          & 88.35          & 89.89          & 91.47          & 93.64          & 94.36          \\
\multicolumn{1}{|l|}{\text{RelaxedSimilarity(P2)}}      & 85.57          & 88.28          & 90.03          & 91.98          & 92.92          & 93.9           \\
\multicolumn{1}{|l|}{\text{InverseKernel(P1+Init+P2)}}  & 87.03          & 89.79          & 90.94          & 92.59          & 93.83          & 94.58          \\ 
\multicolumn{1}{|l|}{\text{GaussianKernel(P2)}}         & 86.24          & 88.91          & 90.76          & 92.69          & 93.56          & 94.53          \\ 
\multicolumn{1}{|l|}{\bftab{GaussianKernel(P1+Init+P2)}} & \bftab{89.03} & \bftab{89.92} & \bftab{92.08} & \bftab{93.46} & \bftab{94.21} & \bftab{94.6}  \\ \hline
\end{tabular}

\label{tab:resutls-resnet-c10}
\end{table}

\begin{table}[h]
\caption{Results with ResNet50 on CIFAR-100.}
\centering
\begin{tabular}{|c|c|c|c|c|c|c|}
\cline{2-7} \hline
   \textbf{Training}                                              & \textbf{10\%}  & \textbf{20\%}  & \textbf{30\%}  & \textbf{50\%}  & \textbf{75\%}  & \textbf{100\%} \\ \hline
\multicolumn{1}{|l|}{\text{SoftMax(P1)}}                & 35.59          & 45.63          & 62.72          & 69.00          & 73.05          & 76.57          \\ 
\multicolumn{1}{|l|}{\text{RelaxedSimilarity(P2)}}      & \bftab{50.91}          & 61.35          & 66.17          & 70.82          & 73.83          & 76.28          \\ 
\multicolumn{1}{|l|}{\text{InverseKernel(P1+Init+P2)}}  & 47.47          & 61.73          & 66.37          & 70.48          & 73.59          & \bftab{76.82} \\ 
\multicolumn{1}{|l|}{\text{GaussianKernel(P2)}}         & 49.54          & 60.98          & 65.96          & 70.48          & 74.01          & 76.21          \\ 
\multicolumn{1}{|l|}{\text{GaussianKernel(P1+Init+P2)}} & {50.56} & \bftab{65.77} & \bftab{66.45} & \bftab{73.19} & \bftab{75.35} & 76.76 \\ \hline
\end{tabular}

\label{tab:results-resnet-c100}
\end{table}

\begin{table}[h]
\caption{Relative improvement in accuracy with respect to a network using SoftMax. The value is shown for different number of centers on CIFAR-10.}
\centering
\begin{tabular}{|c|c|c|c|c|c|c|}
\cline{2-7} \hline
\multirow{1}{*}{\textbf{Training}     }    & \multirow{1}{*}{\textbf{Feature Extractor}}        & \multicolumn{5}{c|}{\textbf{Number of centers}}                                \\ \cline{3-7} 

                                              &  & \textbf{1}  & \textbf{2}  & \textbf{5}  & \textbf{10}  & \textbf{20}   \\ \hline
\multicolumn{1}{|l|}{\text{GaussianKernel(P2)}}                & CNN          &     -7.35      &  -16.77          & -34.93          & -57.01          & -42.37          \\
\multicolumn{1}{|l|}{\text{GaussianKernel(P1+Init+P2)}}      & CNN          &     6.45      &  6.68          & 6.73          & 7
03 & 7.12           \\
\multicolumn{1}{|l|}{\text{GaussianKernel(P2)}}  & ResNet50          &  1.35          &  0.95          &      1.16     & 0.76          & 1.19         \\ 
\multicolumn{1}{|l|}{\text{GaussianKernel(P1+Init+P2)}}         & ResNet50          & 4.64          & 4.70          & 4.75          & 4.9          & 4.78          \\ 
 \hline
\end{tabular}

\label{tab:several-clusters}
\end{table}

\begin{figure}[ht]
\centering
\subfloat[CIFAR-10]{\label{fig:mdleft}{\includegraphics[width=0.5\textwidth]{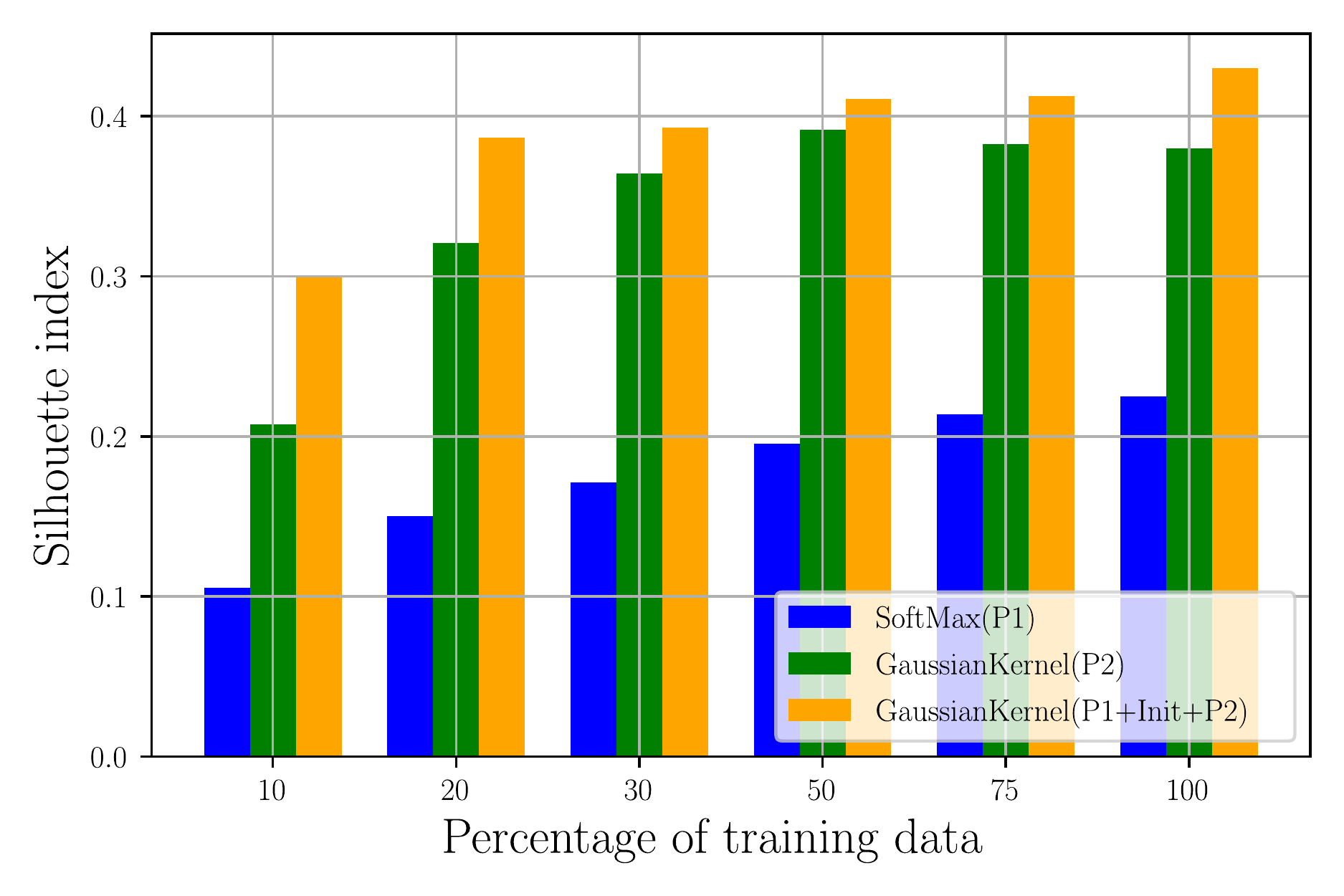}}}\hfill
\subfloat[CIFAR-100]{\label{fig:mdright}{\includegraphics[width=0.5\textwidth]{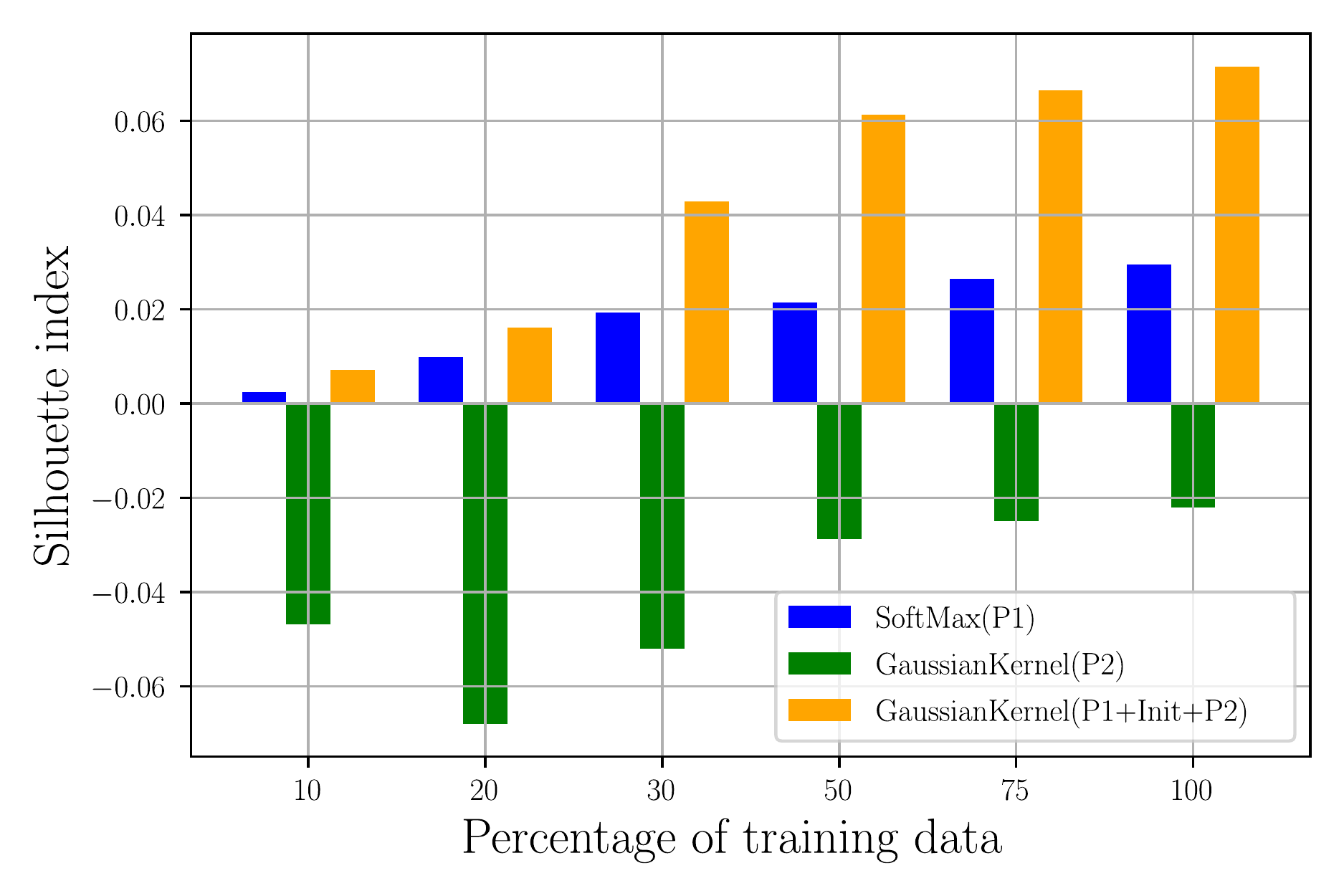}}}

\caption{Silhouette Score for different percentages of training data and with CNN as feature extractor.The neural network trained in two phases and including initialization (\textit{P1+Init+P2}) achieves the best silhouette index for all the percentages in both datasets.}
\label{fig:silhouette}
\end{figure}

\subsection{Goal of the experiments}

Our main experimental objective is to show how procedure specified in Algorithm 1 improves the training of a neural network in a supervised classification task in different data regimes. Therefore, we make comparisons of the accuracy for 10\%, 20\%, 30\%, 50\%, 75\% and 100\% of the training data. We also aim to show that the feature extractors trained with kernelized outputs produce embeddings with higher inter-class separability and intra-class compactness, which explains the performance increase.

\subsection{Experimental setup}

We conduct experiments on CIFAR-10 and CIFAR-100 using two different feature extractors: 1) a simple CNN with two convolutional layers followed by two fully convolutional layers and whose architecture is a variant of the one used in \cite{snell2017prototypical}, and 2) ResNet50, specifically the output of the last average pooling layer. At the outputs of the feature extractor, two fully connected layers are attached, with 512 and 64 neurons respectively. After the output of the last fully connected layer, a batch normalization layer is applied. Finally, the output of the batch normalization is used as input for the last layer, which can be SoftMax or a RBF kernel output. 

The training is stopped after 500 epochs or if there is not improvement after 20 epochs in the validation accuracy. The optimizer is RMSProp with learning rate of 0.00001. Moreover, the batch size is 32 and size of the embeddings, which are the inputs to the output layer, is 64.

\begin{figure}[ht]
\centering
\subfloat[CIFAR-10]{\label{fig:mdleft}{\includegraphics[width=0.5\textwidth]{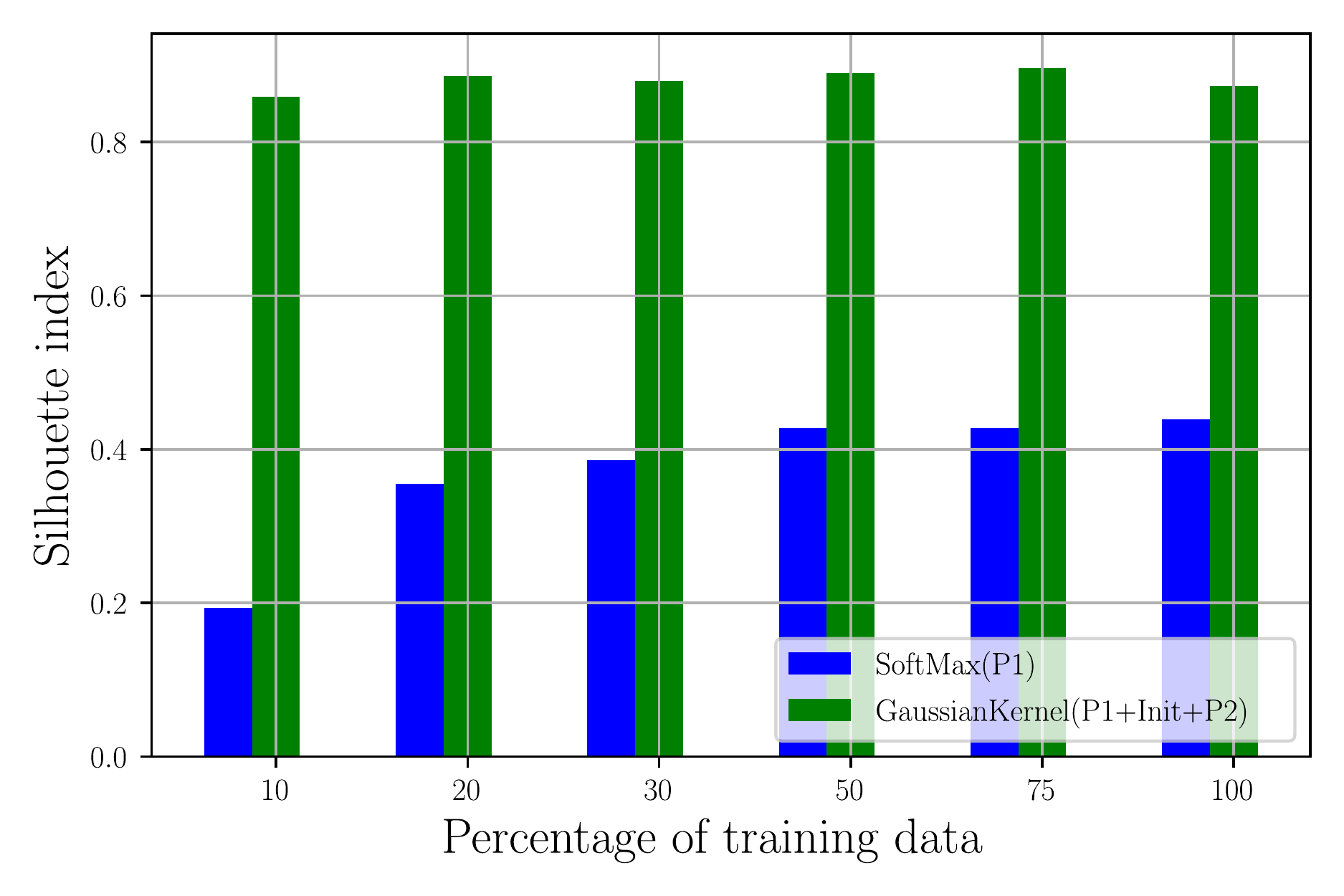}}}\hfill
\subfloat[CIFAR-100]{\label{fig:mdright}{\includegraphics[width=0.5\textwidth]{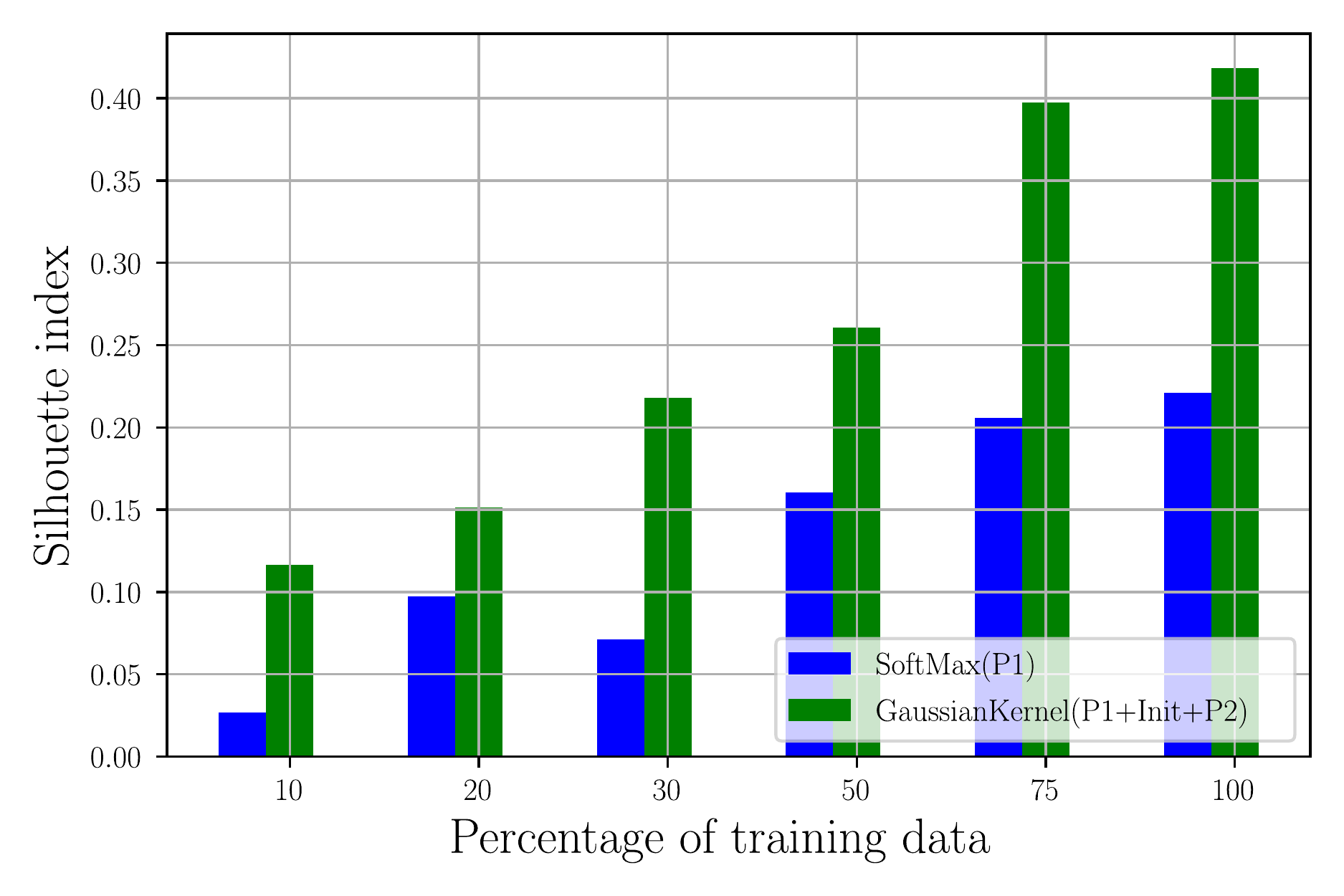}}}

\caption{Silhouette Score for different percentages of training data and with ResNet50 as feature extractor. The neural network trained in two phases and including initialization (\textit{P1+Init+P2}) achieves the best silhouette index for all the percentages in both datasets.}
\label{fig:silhouette}
\end{figure}

\subsection{Performance comparison for different training percentages}

To measure how the phase 1 and the initialization improve the accuracy on test, we compare the results on different tables. In the tables 1-4, the results of the performance in test for different training settings are shown. It is also important to note that the percentage of test data is the same, although the percentage of training data changes. To reference the training procedure, we include the following notation: 

\begin{itemize}
    \item \textit{SoftMax(P1)} refers a network trained using only SoftMax as output.
    \item \textit{InverseKernel(P2)} refers to a network trained only using InverseKernel as output. The same holds respectively for \textit{GaussianKernel(P2)}.
    \item \textit{RelaxedSimilarity(P2)} refers to the output layers proposed in \cite{Qian} which does not have any initialization.
    \item \textit{InverseKernel(P1+P2)} refers to a network trained applying phase 1 and phase 2 from Algorithm 1, but without initializing.
    \item \textit{InverseKernel(P1+Init+P2)} refers to a network trained with all steps defined in Algorithm 1 using InverseKernel in the RPF output. The same applies to \textit{GaussianKernel(P1+Init+P2)}.
    \item \textit{InverseKernel(P1+Init2+P2)} refers to a network trained with all steps defined in Algorithm 1 using InverseKernel as output layer, but using \textit{K-Medoids} instead of \textit{K-Means} as clustering algorithm to get the centroids.
    
\end{itemize}

The results with the CNN network corresponds to the average of three runs, whereas the results on the ResNet50 are the result of only one run. For all the experiments, only one center per class ($|K_c|=1$) is used. This allows fairness in the comparisons by avoiding more complex models. Besides the pre-training phases mentioned here, it is important to mention that the ResNet50 used in all experiments was pre-trained on ImageNet. 

As seen in the tables 1-4, the model with Gaussian kernel in the output layer and trained through the two phases with \textit{K-Means} initialization achieved always the best performance for different percentages of training data.  Interestingly, this model performs always better than \textit{SoftMax(P1)} and \textit{GaussianKernel(P2)}. Also the two-phase algorithm improves the final accuracy of the networks using inverse kernel in all the experiments. Additionally, The initialization procedure (\textit{Init}) also very important, as the accuracy decreases dramatically by using \textit{K-Medoids} instead of \textit{K-Means} in \textit{InverseKernel(P1+Init2+P2)}. This may happen, since \textit{K-Medoids} finds centers that minimize a distortion measure (equation \ref{Distortion}), which is related to the loss function, but setting the condition that the centers should be one of the samples. On the other hand, \textit{K-Means} does not set this constraint. \\

By looking closely to the results in figure \ref{fig:accuracy-percentages}, it is possible to notice that the performance difference over traditional SoftMax is higher specially in lower percentages of training data (10\%-50\%) with ResNet50 as feature extractor in both CIFAR-10 and CIFAR-100. 

\subsection{Performance with different numbers of centers per class}

The equations \ref{eq:7} and \ref{eq:8} represent the output of a RBF output layer for classification and is mathematically equivalent to kernel regression and also generalize the possibility of having more than one center per class. The table \ref{tab:several-clusters} shows the performance improvement when using several number of centers per class. When using the two phases training, the accuracy slightly raises as the number of centers increases. Otherwise, the training with more centers exhibits a bad or even worse performance.\\

\begin{figure}[ht]
\centering
\subfloat[With two-phase training]{\label{fig:heatmapa}{\includegraphics[width=0.50\textwidth]{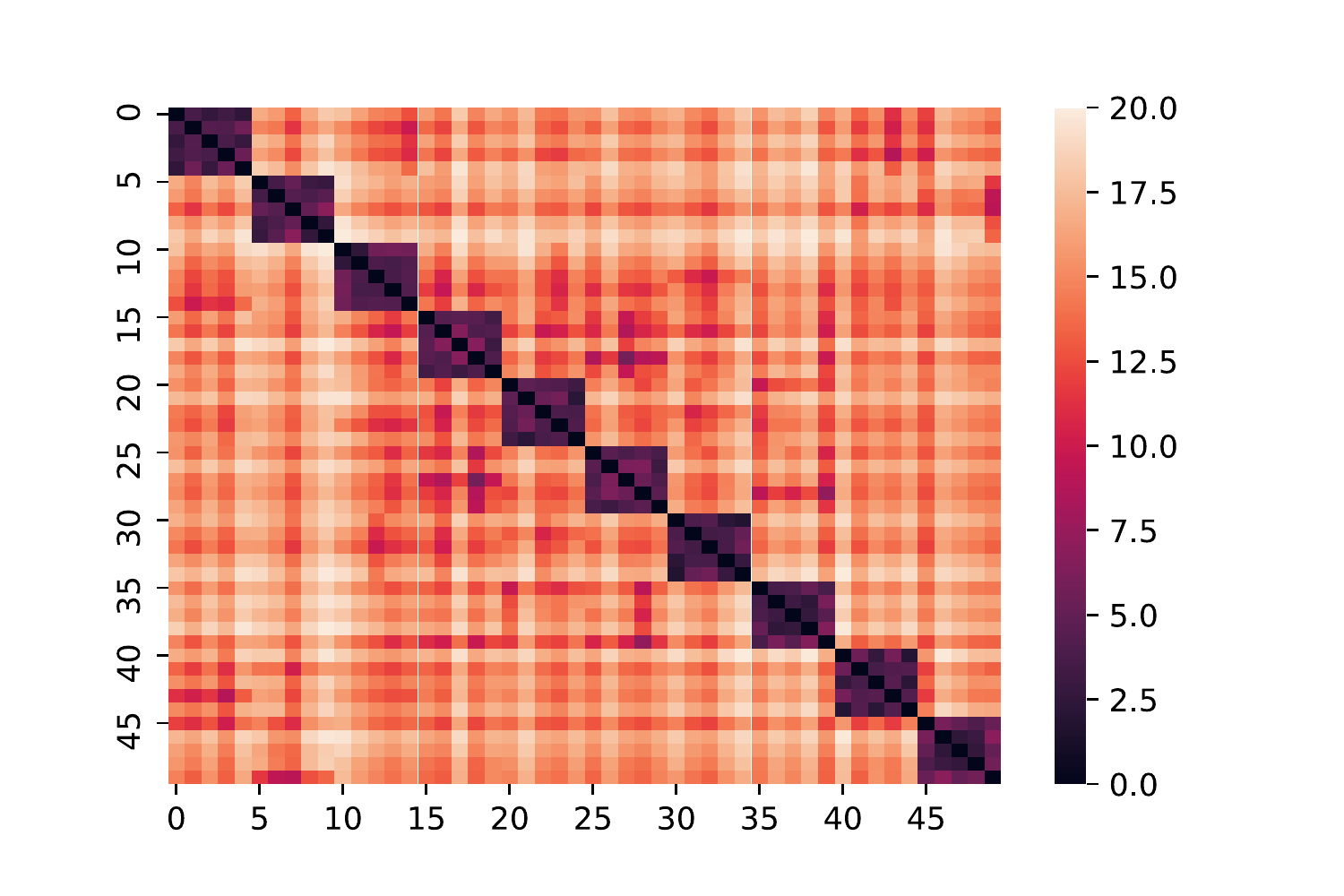}}} \hfill
\subfloat[Without two-phase training]{\label{fig:heatmapb}{\includegraphics[width=0.50\textwidth]{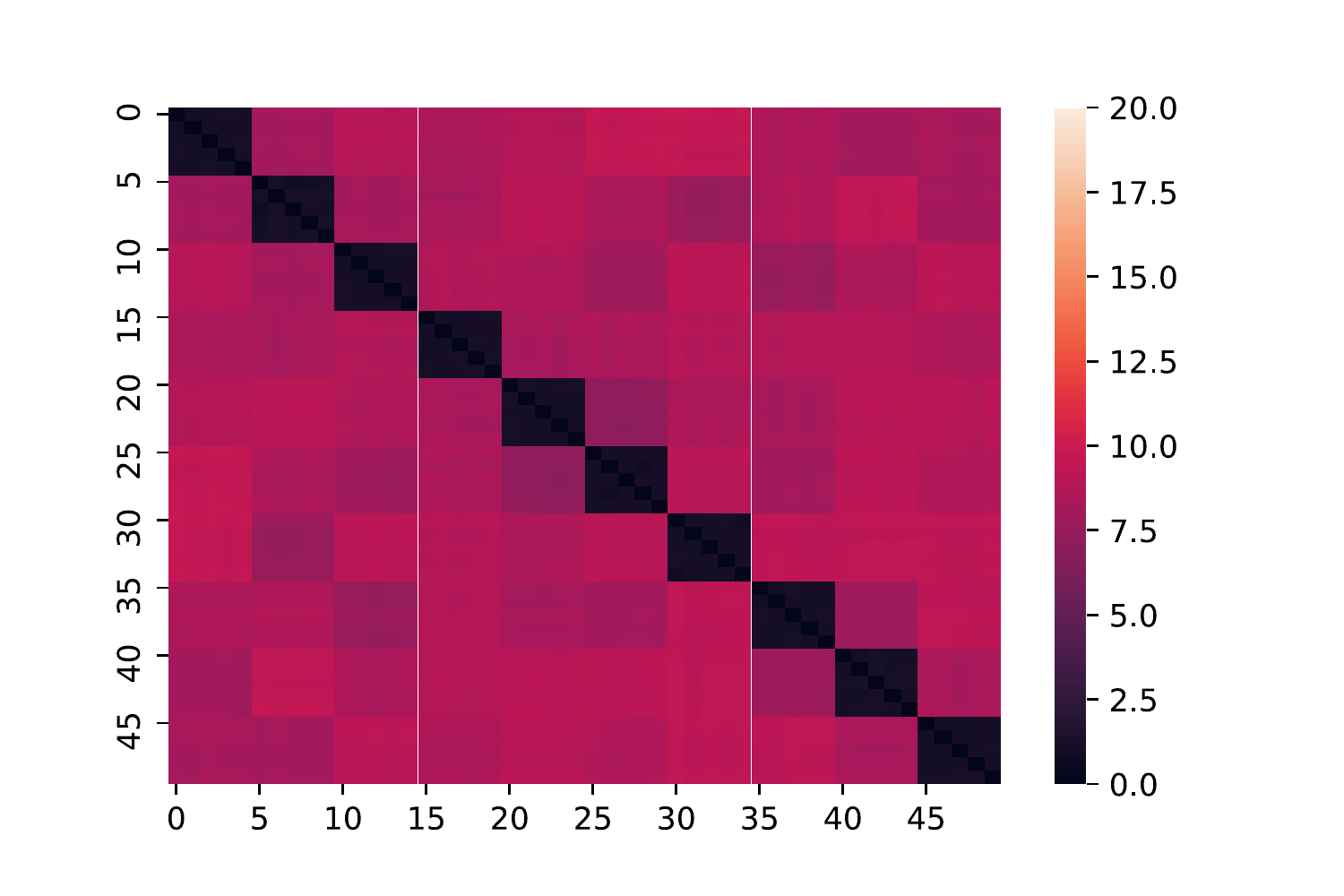}}}

\caption{Heatmap of the euclidean distances between the final trained centers for CIFAR-10 using ResNet50 with five centers per class in the output layer. When two-phase training is not used, there is a "collapse" in the centers, where they fall very close each other.}
\label{fig:heatmap}
\end{figure}

By examining the figure \ref{fig:heatmapa}, it is possible to determine that through the two-phase training, we get centers which are more separated, even when they are from the same class. On the other hand, when there is only a traditional training of the RBF kernel output (figure \ref{fig:heatmapb}), the final centers of all the classes fall very close each other.   Moreover, the centers belonging to the same class tend to be similar. 

\subsection{Measuring the compactness and separability of the classes in the embedding space}

We explore how the compactness and the separability in the embedding space is affected by using the two-phase training. Hereby, compactness refers to the intra-class distance: how close are the embeddings of the samples belonging to the same class. On the other hand, separability refers to the inter-class distance, in other words, how far are the embeddings of samples belonging to the same class. In both cases, the membership of a sample to one class is determined by the trained network, not by the original label. Similarly as in clustering settings,  embedding spaces that enable compact and well-separated classes are desirable, since they enable higher performance and robustness. 

To assess the quality of the embedding spaces when having one center per class, we use the \textit{Silhouette index} \cite{Rousseeuw}, which trade off the compactness and separability of the classes in the embedding space and whose values are between -1 and 1. The closer to 1, the better is the compactness and separability ratio. As shown in figure \ref{fig:silhouette}, the RBF output networks with two-phase training achieve the higher \textit{Silhouette index} than merely using SoftMax or Gaussian Kernel with traditional training. In fact, the \textit{Silhouette index} computed on embeddings of the CIFAR-100 dataset is lower than 0, which indicate a bad quality embedding. 

\begin{figure}[h!] 
\centering
\subfloat[SofMax(P1)]{\label{fig:tsnea}{\includegraphics[width=0.4\textwidth]{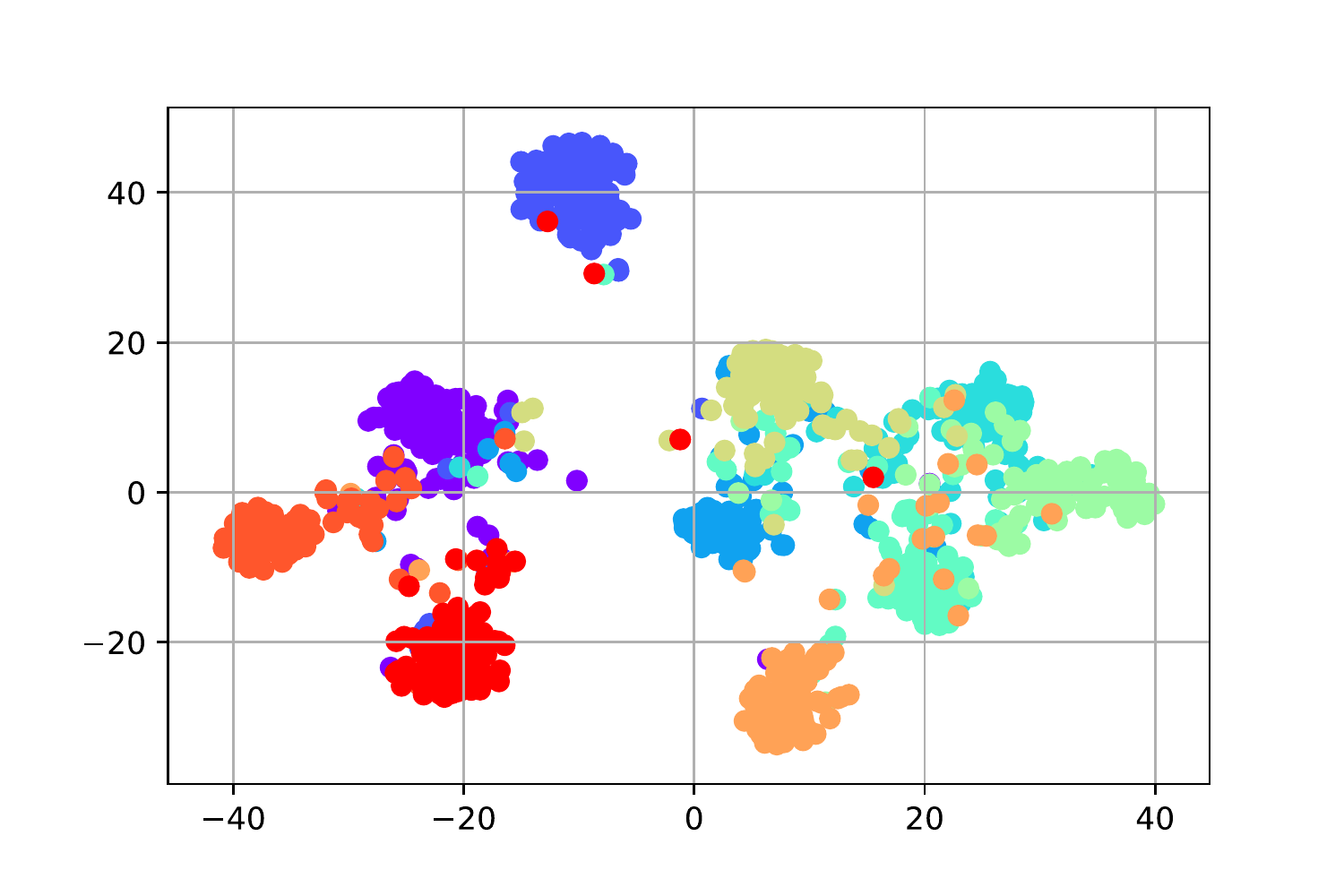}}}\vspace{0.00mm} 
\subfloat[Gaussian Kernel(P2)]{\label{fig:tsneb}{\includegraphics[width=0.4\textwidth]{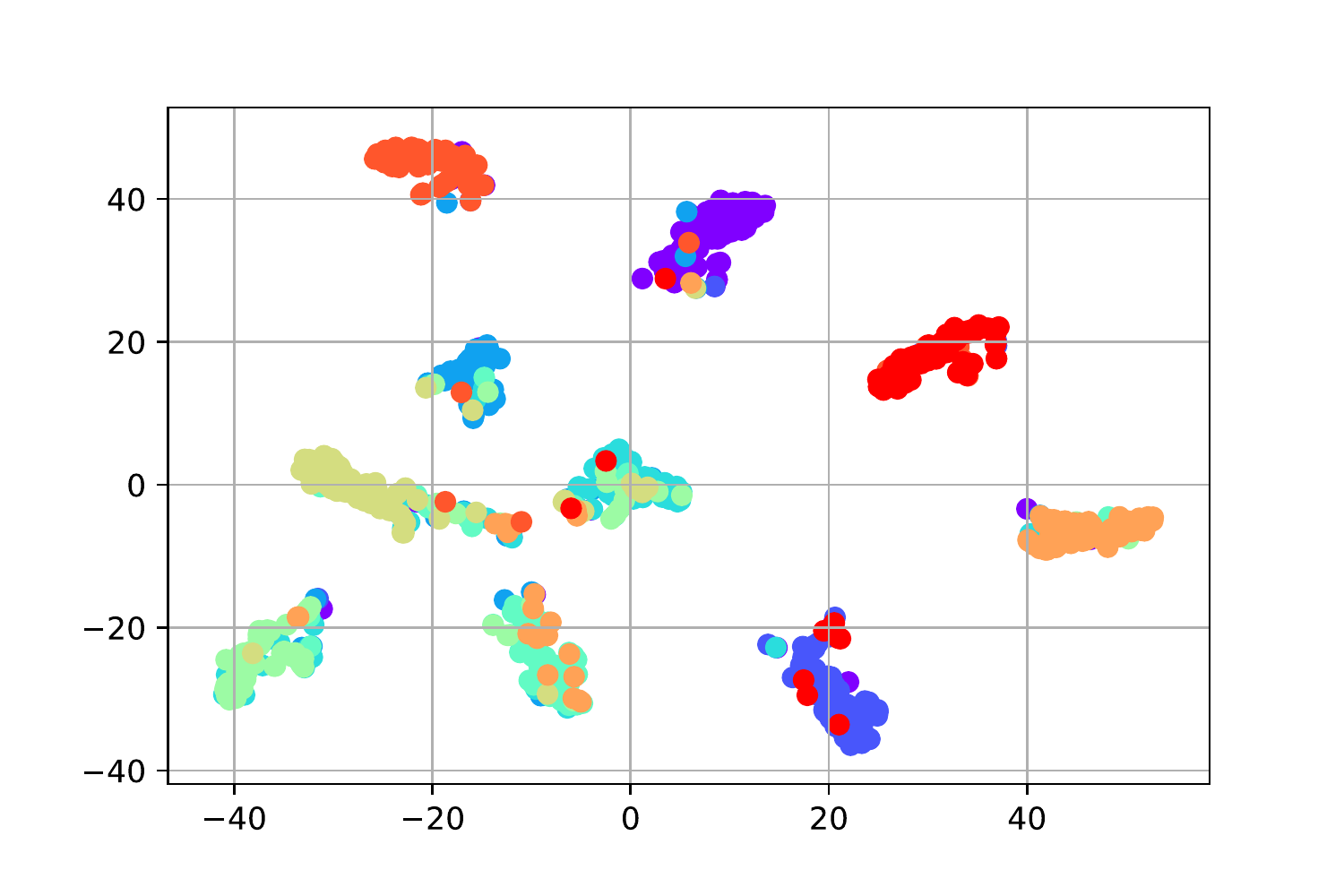}}}\vspace{0.00mm}
\subfloat[Gaussian Kernel(P1+Init+P2)]{\label{fig:tsnec}{\includegraphics[width=0.4\textwidth]{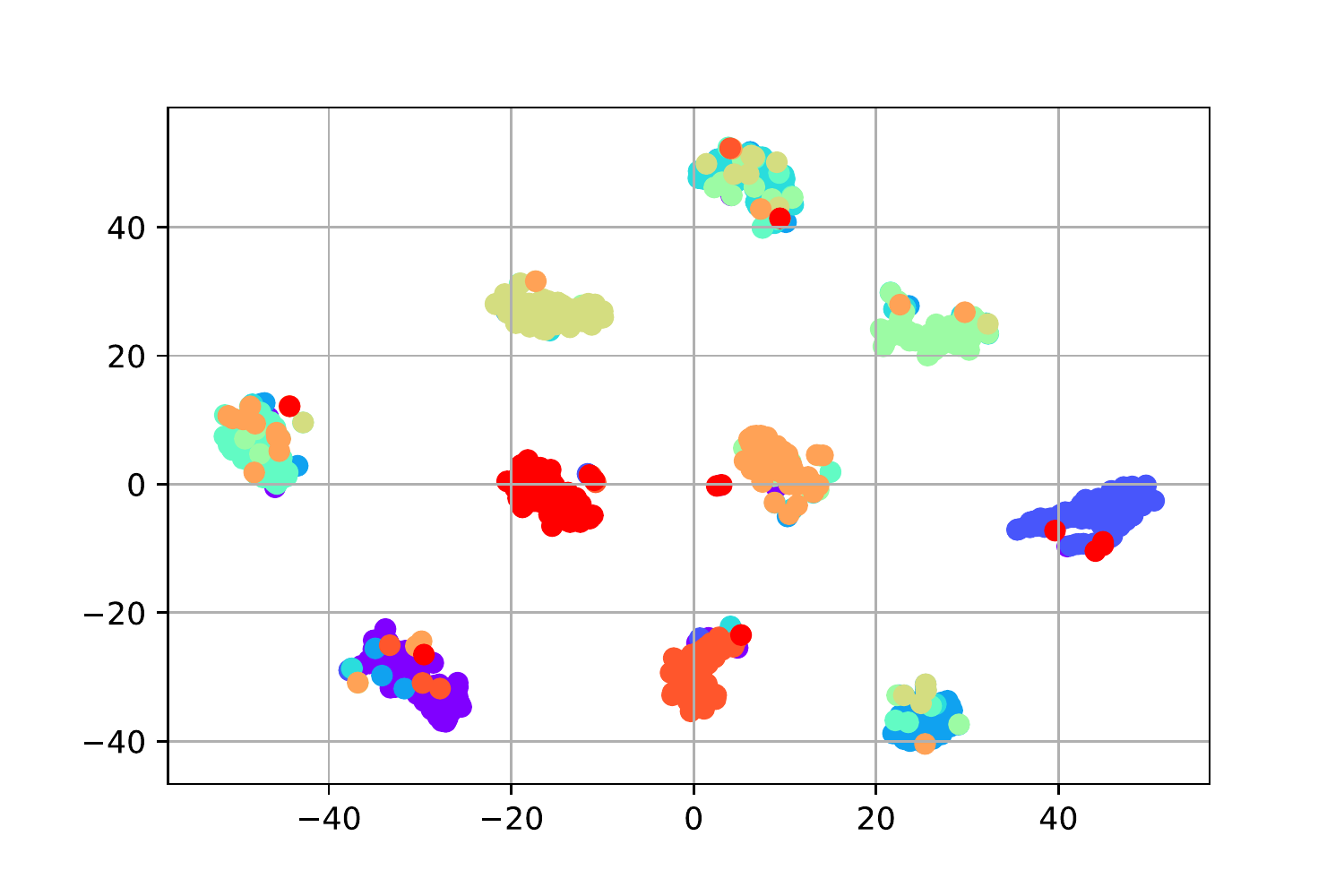}}}

\caption{T-SNE of the test sample embeddings obtained after training with different output layers. The RBF with two-phase training has more compact and better separated classes. }
\label{tsne}
\end{figure}

It is possible to perceive the compactness and separability achieve after the training in the Figure \ref{tsne}. The plots represent the TSNE projections of the test sample embeddings. As seen, the two-phase training (\ref{fig:tsnec}) achieves the best compactness and separability. Additionally, the SoftMax (\ref{fig:tsnea}) still separate the class, but without a high compactness. Training a RBF network without the two-phases also yields compact and well-separated classes (\ref{fig:tsneb}).
For several centers, we use the distortion measure over the test dataset in order to measure the compactness. The distortion measure is $\mathcal{D}_{Test}$ in the equation \ref{Distortion}, where the minimal the distortion is, the more compact are the classes. The lowest distortion, therefore the highest compactness, is achieved when using the two-phase training.

\begin{equation} \label{Distortion}
    Distortion = \sum_{x_i \in \mathcal{D}_{Test}} \mathrm{min}_{\boldsymbol{\mu}_j \in K_{y_i}}||f_{\theta}(x_i)-\boldsymbol{\mu}_j||_2^2
\end{equation}

The measures of the distortion of the embeddings with respect to the final prototypes are shown in the table \ref{DistortionTable} for different numbers of centers. It can be seen that with the two-phase training the final embeddings also achieve better compactness. 

\begin{table}[]
\centering
\caption{Distortion for different number of centers for ResNet50 on CIFAR-10}
\begin{tabular}{|c|C{10mm}|C{10mm}|C{10mm}|}
\hline
\multirow{1}{*}{\textbf{Training}} & \multicolumn{3}{c|}{\textbf{Number of centers} }\\ \cline{2-4} 
                          & 5              & 10             & 20             \\ \hline
GaussianKernel(P1+Init+P2)                  & 5372           & 3732           & 5233           \\ \hline
GaussianKernel(P2)                   & 9950           & 6347           & 6297           \\ \hline
\end{tabular}
\label{DistortionTable}
\end{table}

\subsection{Effect of the initialization}

The main objective of Phase 1 and the subsequent initialization is to find initial centers that are reasonably well-localized. Hence, the embeddings generate a significant activation in a small fraction of the basis function, which enables faster and more effective learning when training the RBF network in Phase 2 \cite{Bishop}. Additionally, the proposed initialization for the RBF layer can be seen as a regularization due to the higher performance improvement in low data regimes. The equation \ref{eq:9} shows how, when having one center per class, the normalized RBF can be parametrized as a linear model where the bias ($b_c$) is a term that penalizes the magnitude of the centers.

\section{Conclusion and future work} \label{Conclusion}

Our experiments show that by following the proposed two-phase training on networks with RBF kernel output, it is possible to improve the accuracy with respect to the same network without any kind of pre-training, and also, respect a network with SoftMax layer as output. This improvement is specially higher when having less training data and deep networks (i.e. ResNet50). One of the reasons for this behaviour lays on the compact and well-separated embeddings that the deep networks  as feature extractors find by using RBF kernel output and being trained with two-phases. In fact, we have empirically shown that RBF kernel outputs enables embeddings spaces with a better balance between higher inter-class distance and lower intra-class. Similarly, when having several centers per class, these training allows the RBF networks to find embedding spaces where clusters from the same class are more diverse while maintaining a separation with respect to the centers of other classes. 

For future research, it would be interesting to explore the robustness of the models after two-phase method. Due to the compactness and well-separability, these models may exhibit good performance in adversarial attacks. Another potential improvement is to explore the performance of the models after two-phase training, where the phase 1 did not fully converge.  Finally, another possible promising direction is to adapt the algorithm to work with Gaussian mixtures models in the output layer.


\end{document}